\documentclass[twoside,american]{elsarticle}
\usepackage[T1]{fontenc}
\usepackage[utf8]{inputenc}
\usepackage{float}
\usepackage{amsmath}
\usepackage{amsthm}
\usepackage{amssymb}
\usepackage{graphicx}

\makeatletter
\theoremstyle{plain}
\newtheorem{thm}{\protect\theoremname}
\theoremstyle{definition}
\newtheorem{example}[thm]{\protect\examplename}
\theoremstyle{remark}
\newtheorem{rem}[thm]{\protect\remarkname}
\ifx\proof\undefined
\newenvironment{proof}[1][\protect\proofname]{\par
	\normalfont\topsep6\p@\@plus6\p@\relax
	\trivlist
	\itemindent\parindent
	\item[\hskip\labelsep\scshape #1]\ignorespaces
}{%
	\endtrivlist\@endpefalse
}
\providecommand{\proofname}{Proof}
\fi

\journal{Elsevier}

\ifdefined\showcaptionsetup
 \PassOptionsToPackage{caption=false}{subfig}
\fi
\usepackage{subfig}
\makeatother

\usepackage{babel}
\providecommand{\examplename}{Example}
\providecommand{\remarkname}{Remark}
\providecommand{\theoremname}{Theorem}

\begin{document}
\begin{frontmatter}
\title{\textbf{Establishing a leader in a pairwise comparisons method}}
\author[wms]{Jacek Szybowski}
\ead{szybowsk@agh.edu.pl}
\author[kis]{Konrad Kułakowski\corref{cor1}}
\ead{kkulak@agh.edu.pl}
\author[sa]{Jiri Mazurek}
\ead{szybowsk@agh.edu.pl}
\author[kis]{Sebastian Ernst}
\ead{ernst@agh.edu.pl}
\cortext[cor1]{Corresponding author}
\address[wms]{AGH University of Krakow, The Faculty of Applied Mathematics, al.
A. Mickiewicza 30, 30-059 Krakow, Poland}
\address[kis]{AGH University of Krakow, The Department of Applied Computer Science,
al. A. Mickiewicza 30, 30-059 Krakow, Poland}
\address[sa]{Silesian University in Opava, School of Business Administration in
Karvina, Univerzitní nám. 1934, 733 40 Karviná, Czech Republic}
\begin{abstract}
Like electoral systems, decision-making methods are also vulnerable
to manipulation by decision-makers. The ability to effectively defend
against such threats can only come from thoroughly understanding the
manipulation mechanisms. In the presented article, we show two algorithms
that can be used to launch a manipulation attack. They allow for equating
the weights of two selected alternatives in the pairwise comparison
method and, consequently, choosing a leader. The theoretical considerations
are accompanied by a Monte Carlo simulation showing the relationship
between the size of the PC matrix, the degree of inconsistency, and
the ease of manipulation. This work is a~continuation of our previous
research published in the paper \citep{Szybowski2023aomo}.
\end{abstract}
\begin{keyword}
pairwise comparisons \sep data manipulation \sep rank reversal \sep
orthogonal projections
\end{keyword}
\end{frontmatter}

\section{Introduction}

The pairwise comparisons method (PC) constitutes a convenient and
broadly applied tool for a complexity reduction in the multiple criteria
decision-making (MCDM) frameworks such as the Analytic Hierarchy Process
(AHP) \citep{Saaty1977asmf}, Best-Worst Method (BWM) \citep{Rezaei2015bwmc},
MACBETH \citep{BanaECosta2016otmf}, or PROMETHEE \citep{Brans2016pm}.

In recent decades, many researchers studied PC methods intensively
concerning its consistency, optimal derivation of a priority vector,
priority vector’s desirable properties, and other aspects, see e.g.
\citep{Csato2020otmo,Kulakowski2015otpo,Kulakowski2021otsb,Mazurek2023aipc}.
Since the objective of a PC method is to rank compared objects (usually
alternatives or criteria) from the best to the worst, it may happen
that an expert deliberately distorts one or more pairwise comparisons
to promote a selected object, see e.g. \citep{Kulakowski2024rhao,Yager2001pspm,Yager2002dasm}.
In particular, the problem of preference manipulation has gained attention
in the context of group decision-making (see, e.g. \citep{Dong2018swmi,Dong2021cras,Lev2019rgmi,Liang2024acmc,Sasaki2023smig}),
or electoral systems analysis (\citep{Brandt2016hocs,Faliszewski2010uctp,Gibbard1973movs}).
The studies above focused on manipulation detection, various anti-manipulation
strategies (mainly through some penalization brought upon a manipulator),
or an estimation of manipulation robustness. Prevention of manipulation
was discussed, e.g., in \citep{Kulakowski2024rhao,Sun2022aatp,Wu2021aofm}.

In particular, a recent study by Kułakowski et al. \citep{Kulakowski2024rhao}
introduced two heuristics enabling the detection of manipulators and
minimizing their effect on the group consensus by diminishing their
weights. The first heuristic is based on the assumption that manipulators
will provide judgments that can be considered outliers concerning
those of the other experts in the group. The second heuristic assumes
dishonest judgments are less consistent than the average consistency
of the group.

The presented study is a follow-up of the work by Szybowski et al.
\citep{Szybowski2023aomo}, where an algorithm balancing the weights
of two given alternatives of a pairwise comparisons matrix (\emph{EQ
algorithm}) has been introduced. This study aims to introduce a modification
of the EQ algorithm that is more efficient in the case of its multiple
uses and to propose two other algorithms based on the EQ algorithm
(\emph{greedy and bubble sort}) capable of altering the best alternative
by a minimal change in elements of an original additive PC matrix.
Further, we define the so-called \emph{Average Ranking Stability Index}
(ARSI) as a measurement of ranking manipulation's difficulty. Last
but not least, we perform Monte Carlo simulations to analyze relationships
between the size of a PC matrix, its inconsistency, and the degree
of manipulation difficulty given by the ARSI. In the proposed method,
we use PC matrix orthogonalization. We can also use this technique
in procedures to increase the consistency of PC matrices \citep{Benitez2024ceof,Koczkodaj2020oopo}.

The paper is organized as follows: Section \ref{sec:Preliminaries}
provides preliminaries, Section \ref{sec:Establishing-a-leading}
presents new algorithms, and Section \ref{sec:Monte-Carlo-Simulation}
includes numerical (Monte Carlo) simulations. Conclusions close the
article.

\section{Preliminaries\protect\label{sec:Preliminaries}}

\subsection{Multiplicative and additive pairwise comparisons systems}

Let $E=\{e_{1},\ldots,e_{n}\}$ be a finite set of alternatives, $n\geq2$,
and the goal is to rank all alternatives from the best to the worst
by pairwise comparisons.
\begin{itemize}
\par 
\item In the multiplicative pairwise comparisons (MPCs) framework, an expert
expresses his/her judgment of a relative preference (importance) of
$e_{i}$ and $e_{j}$ by the value $m_{ij}\in\mathbb{R^{+}}$, where
$m_{ij}>1$ means $e_{i}$ is preferred over $e_{j}$, and $m_{ij}=1$
denotes equal preference of both alternatives.

\bigskip{}

MPCs are \textit{reciprocal}, if:

$m_{ij}=1/m_{ji};\forall i,j\in\lbrace1,...,n\rbrace$.

MPCs are \textit{consistent}, if:

$m_{ij}\cdot m_{jk}=m_{ik};\forall i,j\in\lbrace1,...,n\rbrace$.

All MPCs are conveniently arranged into an $n\times n$ multiplicative
pairwise comparisons matrix $M=[m_{ij}]$, and a priority vector (vector
of alternatives' weights) is then calculated by the eigenvector \citep{Saaty1977asmf}
or the (row) geometric mean method \citep{Crawford1987tgmp}.

Inconsistency of an MPC matrix $M$ can be estimated by the consistency
index ($CI$) \citep{Saaty1977asmf}:

$CI(M)=\dfrac{\lambda_{max}-n}{n-1}$, where $\lambda_{max}$ denotes
the maximal eigenvalue of $M$. Of course, there are a number of other
methods for determining the degree of inconsistency of a PC matrix
such as the Koczkodaj's index \citep{Koczkodaj2018aoii}, Kazibudzki's
Square Logarithm Deviations index \citep{Kazibudzki2022oeop} or Barzilai's
error \citep{Barzilai1998cmfp}. A comprehensive review of methods
for measuring inconsistency in PC matrices can be found in \citep{Brunelli2018aaoi}.
In addition to the inconsistency of the PC matrix, the incompleteness
index can also be determined \citep{Kulakowski2019tqoi}. 

\bigskip{}

\item In the additive pairwise comparisons (APCs) framework, an expert expresses
his/her judgment of a relative preference (importance) of $e_{i}$
and $e_{j}$ by the value $a_{ij}\in\mathbb{R}$, where $a_{ij}>0$
means $e_{i}$ is preferred over $e_{j}$, and $a_{ij}=0$ denotes
equal preference of both alternatives.

\bigskip{}

APCs are \textit{reciprocal}, if:

$a_{ij}=-a_{ji};\forall i,j\in\lbrace1,...,n\rbrace$.

APCs are \textit{consistent}, if:

$a_{ij}+a_{jk}=a_{ik};\forall i,j\in\lbrace1,...,n\rbrace$.

All APCs are conveniently arranged into an $n\times n$ additive pairwise
comparisons matrix $A=[a_{ij}]$, and a priority vector (vector of
alternatives' weights) is then calculated by the row arithmetic mean
method \citep{Barzilai1990dwfp}.
\end{itemize}
\bigskip{}

Multiplicative and additive pairwise comparisons share the same group
structure (are isomorphic) \citep{Cavallo2023acso} and can be easily
converted into each other by exponential and logarithmic transformations
respectively:

\bigskip{}

$a_{ij}=log(m_{ij}),m_{ij}=exp(a_{ij})$

\bigskip{}

Both MPC and APC systems have they advantages. While the MPCs are
based on ratios, which are natural to human thinking, APCs enable
to use rich mathematical apparatus of linear algebra, which is especially
convenient for theoretical considerations \citep{Fedrizzi2020tlao}.

The space 
\[
\mathcal{A}:=\{[a_{ij}]:\ \forall i,j\in\{1,\ldots,n\}\ a_{ij}\in\mathbb{R}\textnormal{ and }a_{ij}+a_{ji}=0\},
\]
is a linear space of additive pairwise comparisons matrices (PCMs).
Recall that any linear space is endowed with a (orthogonal) \textit{basis}
and that for two given $n\times n$ matrices $A$ and $B$ their standard
Frobenius product is defined as follows:

\[
\langle A,B\rangle=\sum_{k=1}^{n}\sum_{l=1}^{n}a_{kl}b_{kl},
\]
which induces the Frobenius norm 
\[
||A||=\sqrt{\langle A,A\rangle}
\]
and the Frobenius distance 
\[
d(A,B)=||A-B||.
\]

\subsection{Ranking stability index}

In the additive pairwise comparisons method it is usually assumed
that the elements of a PCM fall within a certain range $[-M,M]$,
for a fixed $M>0$. In this case, according to \citep{Szybowski2023aomo}
the Ranking Stability Index of alternatives $e_{i}$ and $e_{j}$
has been defined as 
\[
RSI_{ij}^{M}=\frac{|\sum_{k=1}^{n}(a_{ik}-a_{jk})|}{2M}.
\]
This index expresses a rescaled distance of the weights of the $i$-th
and $j$-th alternatives.

The Ranking Stability Index for a PCM $A$ is given by the formula
\[
RSI^{M}(A)=\min_{1\leq i\leq j\leq n}RSI_{ij}^{M},
\]
and it measures the ease of the easiest manipulation.

However, sometimes the decision process is more complicated and some
attempts of manipulations may not be that obvious. Therefore, it could
be useful to define the Average Ranking Stability Index for $A$ as
follows: 
\[
ARSI^{M}(A)=\frac{2}{n(n-1)^{2}}\sum_{1\leq i\leq j\leq n}RSI_{ij}^{M}.
\]

Since for all $i,j$ 
\[
0\leq RSI_{ij}^{M}\leq n-1,
\]
we immediately get 
\[
0\leq ARSI^{M}\leq1.
\]

\section{Establishing a leading alternative\protect\label{sec:Establishing-a-leading}}

\subsection{Equating two alternatives}

Let us recall the algorithm of finding the best approximation of a
given PCM $A$, which equates the weights of two given alternatives
$e_{i}$ and $e_{j}$ (for $i<j$). This algorithm has been introduced
in \citep{Szybowski2023aomo} and we will denote it by EQ($A,i,j$).

In the beginning, we consider the case $i=1$ and $j=2$.

For this purpose we define: 
\begin{itemize}
\item the tie space ${\cal A}_{12}$, i.e. the $\frac{n^{2}-n-2}{2}$-dimensional
subspace of all additive PCMs which induce the ranking such that alternatives
$1$ and $2$ are equal:

\[
{\cal A}_{12}=\left\{ A\in{\cal A}:\sum_{k=1}^{n}a_{1k}=\sum_{k=1}^{n}a_{2k}\right\} ,
\]

\item the set 
\[
Z_{12}:=\{(q,r):\ 3\leq q<r\leq n\}.
\]
\end{itemize}
\noindent We define a basis for the tie space ${\cal A}_{12}$ which
consists of additive PCMs $C^{qr}$ ($(q,r)\in Z_{12}$), $E^{1}$,
$F^{p}$ ($p\in\{3,\ldots,n\}$) and $G^{p}$ ($p\in\{3,\ldots,n-1\}$),
whose elements are given by 
\[
c_{kl}^{qr}=\left\{ \begin{array}{rl}
1, & k=q,\ l=r\\
-1, & k=r,\ l=q\\
0, & \textnormal{otherwise}
\end{array}\right..
\]

\[
e_{kl}^{1}=\left\{ \begin{array}{rl}
1, & (k=1,\ l=2)\\
-1, & (k=2,\ l=1)\\
2, & k=2,\ l=n\\
-2, & k=n,\ l=2\\
0, & \textnormal{otherwise}
\end{array}\right.,
\]

\[
f_{kl}^{p}=\left\{ \begin{array}{rl}
1, & (k=1,\ l=p)\textnormal{ or }(k=2,\ l=n)\\
-1, & (k=p,\ l=1)\textnormal{ or }(k=n,\ l=2)\\
0, & \textnormal{otherwise}
\end{array}\right.,
\]

and

\[
g_{kl}^{p}=\left\{ \begin{array}{rl}
1, & (k=2,\ l=p)\textnormal{ or }(k=n,\ l=2)\\
-1, & (k=p,\ l=2)\textnormal{ or }(k=2,\ l=n)\\
0, & \textnormal{otherwise}
\end{array}\right..
\]

\begin{thm}[Theorem 5,\citep{Szybowski2023aomo}]
A family of matrices 
\begin{equation}
{\cal B}=\{B^{p}\}_{p=1}^{\frac{n^{2}-n}{2}-1}:=\{C^{qr}\}_{(q,r)\in Z_{12}}\cup\{E^{1}\}\cup\{F^{p}\}_{p=3}^{n}\cup\{G^{p}\}_{p=3}^{n-1}\label{basis}
\end{equation}
is a basis of ${\cal A}_{12}$. 
\end{thm}
Next, we apply a standard Gram-Schmidt process to the basis 
\[
B^{1},\ldots,B^{\frac{n^{2}-n}{2}-1}
\]
of the vector space ${\cal A}_{12}$ equipped with a standard Frobenius
inner product $\langle\cdot,\cdot\rangle$ and we obtain a pairwise
orthogonal basis 
\begin{equation}
H^{1},\ldots,H^{\frac{n^{2}-n}{2}-1}\label{ort-basis}
\end{equation}
as follows: 
\begin{eqnarray*}
H^{1} & = & B^{1},\\
H^{2} & = & B^{2}-\frac{\langle H^{1},B^{2}\rangle}{\langle H^{1},H^{1}\rangle}H^{1},\\
H^{3} & = & B^{3}-\frac{\langle H^{1},B^{3}\rangle}{\langle H^{1},H^{1}\rangle}H^{1}-\frac{\langle H^{2},B^{3}\rangle}{\langle H^{2},H^{2}\rangle}H^{2},\\
\cdots & = & \cdots\\
H^{\frac{n^{2}-n}{2}-1} & = & B^{\frac{n^{2}-n}{2}-1}-\sum_{p=1}^{\frac{n^{2}-n}{2}-2}\frac{\langle H^{p},B^{\frac{n^{2}-n}{2}-1}\rangle}{\langle H^{p},H^{p}\rangle}H^{p}.
\end{eqnarray*}

\begin{example}
\label{basis4} Consider $n=4$. Then the dimension of ${\cal A}_{12}$
is $\frac{n^{2}-n-2}{2}=5$.

\noindent Since $Z_{12}=\{(3,4)\}$, we get the following basis of
${\cal A}_{12}$:\\

\noindent$B^{1}=C^{34}=\left(\begin{array}{cccc}
0 & 0 & 0 & 0\\
0 & 0 & 0 & 0\\
0 & 0 & 0 & 1\\
0 & 0 & -1 & 0
\end{array}\right),$ $B^{2}=E^{1}=\left(\begin{array}{cccc}
0 & 1 & 0 & 0\\
-1 & 0 & 0 & 2\\
0 & 0 & 0 & 0\\
0 & -2 & 0 & 0
\end{array}\right),$

\noindent$B^{3}=F^{3}=\left(\begin{array}{cccc}
0 & 0 & 1 & 0\\
0 & 0 & 0 & 1\\
-1 & 0 & 0 & 0\\
0 & -1 & 0 & 0
\end{array}\right),$ $B^{4}=F^{4}=\left(\begin{array}{cccc}
0 & 0 & 0 & 1\\
0 & 0 & 0 & 1\\
0 & 0 & 0 & 0\\
-1 & -1 & 0 & 0
\end{array}\right),$

\noindent$B^{5}=G^{3}=\left(\begin{array}{cccc}
0 & 0 & 0 & 0\\
0 & 0 & 1 & -1\\
0 & -1 & 0 & 0\\
0 & 1 & 0 & 0
\end{array}\right).$

Application of the Gram-Schmidt process to this basis results in an
orthogonal basis

\begin{eqnarray*}
H^{1}=B^{1} & = & \left(\begin{array}{cccc}
0 & 0 & 0 & 0\\
0 & 0 & 0 & 0\\
0 & 0 & 0 & 1\\
0 & 0 & -1 & 0
\end{array}\right),\\
\langle H^{1},B^{2}\rangle=0\Rightarrow H^{2}=B^{2} & = & \left(\begin{array}{cccc}
0 & 1 & 0 & 0\\
-1 & 0 & 0 & 2\\
0 & 0 & 0 & 0\\
0 & -2 & 0 & 0
\end{array}\right),\\
\langle H^{1},B^{3}\rangle=0,\ \langle H^{2},B^{3}\rangle=4,\ \langle H^{2},H^{2}\rangle=10 & \Rightarrow\\
\Rightarrow H^{3} & = & \left(\begin{array}{cccc}
0 & -\frac{2}{5} & 1 & 0\\
\frac{2}{5} & 0 & 0 & \frac{1}{5}\\
-1 & 0 & 0 & 0\\
0 & -\frac{1}{5} & 0 & 0
\end{array}\right),\\
\langle H^{1},B^{4}\rangle=0,\ \langle H^{2},B^{4}\rangle=4,\ \langle H^{3},B^{4}\rangle=\frac{2}{5},\ \langle H^{3},H^{3}\rangle=\frac{12}{5} & \Rightarrow\\
\Rightarrow H^{4} & = & \left(\begin{array}{cccc}
0 & -\frac{1}{3} & -\frac{1}{6} & 1\\
\frac{1}{3} & 0 & 0 & \frac{1}{6}\\
\frac{1}{6} & 0 & 0 & 0\\
-1 & -\frac{1}{6} & 0 & 0
\end{array}\right),\\
\langle H^{1},B^{5}\rangle=0,\ \langle H^{2},B^{5}\rangle=-4,\ \langle H^{3},B^{5}\rangle=-\frac{2}{5}\ \langle H^{4},B^{5}\rangle=-\frac{1}{3},\ \\
\langle H^{4},H^{4}\rangle=\frac{7}{3}\Rightarrow H^{5} & = & \left(\begin{array}{cccc}
0 & \frac{2}{7} & \frac{1}{7} & \frac{1}{7}\\
-\frac{2}{7} & 0 & 1 & -\frac{1}{7}\\
-\frac{1}{7} & -1 & 0 & 0\\
-\frac{1}{7} & \frac{1}{7} & 0 & 0
\end{array}\right).
\end{eqnarray*}
\end{example}
Now, for an additive PCM $A$ we find its projection $A'$ onto the
subspace ${\cal A}_{12}$ as a linear combination of the orthogonal
basis vectors 
\[
H^{1},\ldots,H^{\frac{n^{2}-n}{2}-1}:
\]
i.e. 
\[
A'=\varepsilon_{1}H^{1}+\ldots\varepsilon_{\frac{n^{2}-n}{2}-1}H^{\frac{n^{2}-n}{2}-1},
\]
where the factors 
\[
\varepsilon_{1},\ldots,\varepsilon_{\frac{n^{2}-n}{2}-1}
\]
are expressed by formulas:

\[
\varepsilon_{k}=\frac{\langle A,H^{k}\rangle}{\langle H^{k},H^{k}\rangle},\ k=1,\ldots,\frac{n^{2}-n}{2}-1.
\]

Thus, the algorithm EQ($A,1,2$) can be written in a very simple way: 
\begin{enumerate}
\item ${\displaystyle A':=\sum_{k=1}^{\frac{n^{2}-n}{2}-1}\frac{\langle A,H^{k}\rangle}{\langle H^{k},H^{k}\rangle}H^{k};}$ 
\item Return($A'$); 
\end{enumerate}
Now, let us consider the general case, i.e. $1\leq i<j\leq n$. 
\begin{rem}
If $P$ is a matrix of permutation of the $p$-th and $q$-th coordinates,
then 
\[
||PA-PB||=||A-B||,
\]
and 
\[
||AP-BP||=||A-B||,
\]
for each PCMs $A$ and $B$. 
\end{rem}
\begin{proof}
The thesis follows from the fact that we get $P(A-B)$ (and $(A-B)P$)
from $A-B$ by the permutation of the $p$-th and the $q$-th rows
(columns). 
\end{proof}
Thus, in order to find the closest matrix to $A$ equating the $i$-th
and $j$-th alternatives, we first permute alternatives $\{i,j\}$
with $\{1,2\}$, then perform EQ($A,1,2$), and finally permute $\{1,2\}$
with $\{i,j\}$.

Let us define the permutation matrix $P_{ij}=[p_{kl}]_{k,l=1}^{n}$.

If $i=1$ and $j\neq2$, then we put: 
\[
p_{kl}=\left\{ \begin{array}{cl}
1, & (k,l)\in\{(2,j),(j,2)\}\textnormal{ or }k=l\not\in\{2,j\}\\
0, & \textnormal{otherwise,}
\end{array}\right.
\]

If $i\neq1$ and $j=2$, then we put: 
\[
p_{kl}=\left\{ \begin{array}{cl}
1, & (k,l)\in\{(1,i),(i,1)\}\textnormal{ or }k=l\not\in\{1,i\}\\
0, & \textnormal{otherwise,}
\end{array}\right.
\]

If $i\neq1$ and $j\neq2$, then we put: 
\[
p_{kl}=\left\{ \begin{array}{cl}
1, & (k,l)\in\{(1,i),(i,1),(2,j),(j,2)\}\textnormal{ or }k=l\not\in\{1,2,i,j\}\\
0, & \textnormal{otherwise.}
\end{array}\right.
\]

SInce for each $(i,j)\neq(1,2)$ the matrix $P_{ij}$ is orthogonal
we have 
\begin{rem}
$P_{ij}^{-1}=P_{ij}^{T}=P_{ij}$. 
\end{rem}
We are ready to introduce the general algorithm EQ($A,i,j$): 
\begin{enumerate}
\item If $(i,j)\neq(1,2)$, then $A:=P_{ij}AP_{ij}$; 
\item $A$:=EQ($A,1,2$); 
\item If $(i,j)\neq(1,2)$, then $A:=P_{ij}AP_{ij}$; 
\item Return($A$). 
\end{enumerate}
Notice that the above procedure improves the algorithm introduced
in \citet{Szybowski2023aomo}, because:\\
 1. we allow $j=n$,\\
 2. we always use the same orthogonal base $H^{1},\ldots,H^{\frac{n^{2}-n}{2}-1}$
(which is important if we have to run EQ($i,j$) several times for
different $i$ and $j$). 
\begin{thm}[Theorem 9, \citep{Szybowski2023aomo}]
Let $A=[a_{kl}]\in{\cal {A}}$, $i,j\in\{1,\ldots,n\}$, and $A'=[a'_{kl}]$
be the orthogonal projection of $A$ onto ${\cal A}_{ij}$. Then\\
 (1) For each $k\not\in\{i,j\}$ 
\begin{equation}
\sum_{l=1}^{n}a'_{kl}=\sum_{l=1}^{n}a_{kl},\label{eq_weights}
\end{equation}
(2) 
\begin{equation}
\sum_{l=1}^{n}a'_{il}=\sum_{l=1}^{n}a'_{jl}=\frac{\sum_{l=1}^{n}a_{il}+\sum_{l=1}^{n}a_{jl}}{2}.\label{am_weights}
\end{equation}
\end{thm}

\subsection{The algorithm for establishing a leading alternative in a PC method}

Let us present the main algorithm of the paper.

Suppose we have a PCM $A$ and we want to promote the $p$-th alternative
for the first place in the ranking.

\subsubsection{The greedy algorithm\protect\label{subsec:The-greedy-algorithm}}

INPUT: $A,\ p$. 
\begin{enumerate}
\item $q:=$ the number of the best alternative 
\item If $p=q$ then return($A$); 
\item Construct the basis (\ref{basis}); 
\item Apply the Gram-Schmidt process to obtain the basis (\ref{ort-basis}); 
\item repeat\\

\begin{itemize}
\item $A:=$EQ($A,p,q$); 
\item $q:=$ the number of the best alternative; 
\end{itemize}
until ranking($p$) $=$ ranking($q$);\\

\item return($A$); 
\end{enumerate}
\begin{example}
\label{greedy} Let us consider a $4\times4$ ($n=4$) PCM 
\[
A=\left(\begin{array}{cccc}
0 & 1 & 2 & 9\\
-1 & 0 & 1 & 8\\
-2 & -1 & 0 & 7\\
-9 & -8 & -7 & 0
\end{array}\right).
\]

The weights in a ranking vector obtained as the arithmetic means of
elements of rows of $A$ are 
\[
w=(3,2,1,-6)^{T},
\]
so the initial value of $q$ is 1.

Our goal is to promote the fourth alternative ($p=4$) to the first
position in a ranking.

In the example the alternative number 4 is definitely the worst one,
so the algorithm EQ must run $n-1=3$ times, which is the maximal
possible number of iterations.

We construct the basis $B^{1},\ldots,B^{5}$. Next, we apply the Gram-Shmidt
procedure to obtain basis $H^{1},\ldots,H^{5}$. Both bases are described
in Ex. \ref{basis4}.\\

\textbf{THE 1ST ITERATION OF THE LOOP:}\\

We run EQ($A,1,4$), i.e. we calculate:

\[
P_{14}=\left(\begin{array}{cccc}
1 & 0 & 0 & 0\\
0 & 0 & 0 & 1\\
0 & 0 & 1 & 0\\
0 & 1 & 0 & 0
\end{array}\right).
\]

\[
A^{(2)}=P_{14}AP_{14}=\left(\begin{array}{cccc}
0 & 9 & 2 & 1\\
-9 & 0 & -7 & -8\\
-2 & 7 & 0 & -1\\
-1 & 8 & 1 & 0
\end{array}\right).
\]

\[
A^{(3)}=\textnormal{EQ}(A^{(2)},1,2)=\left(\begin{array}{cccc}
0 & 0 & -2.5 & -3.5\\
0 & 0 & -2.5 & -3.5\\
2.5 & 2.5 & 0 & -1\\
3.5 & 3.5 & 1 & 0
\end{array}\right).
\]

\[
A^{(4)}=P_{14}A^{(3)}P_{14}=\left(\begin{array}{cccc}
0 & -3.5 & -2.5 & 0\\
3.5 & 0 & 1 & 3.5\\
2.5 & -1 & 0 & 2.5\\
0 & -3.5 & -2.5 & 0
\end{array}\right).
\]

The ranking vector for $A^{(4)}$ is 
\[
w=(-1.5,2,1,-1.5)^{T},
\]
so the next value of $q$ is 2.\\

\textbf{THE 2ND ITERATION OF THE LOOP:}\\

We run EQ($A^{(4)},2,4$), i.e. we calculate:

\[
P_{24}=\left(\begin{array}{cccc}
0 & 0 & 0 & 1\\
0 & 1 & 0 & 0\\
0 & 0 & 1 & 0\\
1 & 0 & 0 & 0
\end{array}\right).
\]

\[
A^{(5)}=P_{24}A^{(4)}P_{24}=\left(\begin{array}{cccc}
0 & -3.5 & -2.5 & 0\\
3.5 & 0 & 1 & 3.5\\
2.5 & -1 & 0 & 2.5\\
0 & -3.5 & -2.5 & 0
\end{array}\right).
\]

\[
A^{(6)}=\textnormal{EQ}(A^{(5)},1,2)=\left(\begin{array}{cccc}
0 & 0 & -0.75 & 1.75\\
0 & 0 & -0.75 & 1.75\\
0.75 & 0.75 & 0 & 2.5\\
-1.75 & -1.75 & -2.5 & 0
\end{array}\right).
\]

\[
A^{(7)}=P_{24}A^{(6)}P_{24}=\left(\begin{array}{cccc}
0 & -1.75 & -2.5 & -1.75\\
1.75 & 0 & -0.75 & 0\\
2.5 & 0.75 & 0 & 0.75\\
1.75 & 0 & -0.75 & 0
\end{array}\right).
\]

The ranking vector for $A^{(7)}$ is 
\[
w=(-1.5,0.25,1,0.25)^{T},
\]
so the next value of $q$ is 3.\\

\textbf{THE 3RD ITERATION OF THE LOOP:}\\

We run EQ($A^{(7)},3,4$), i.e. we calculate:

\[
P_{34}=\left(\begin{array}{cccc}
0 & 0 & 1 & 0\\
0 & 0 & 0 & 1\\
1 & 0 & 0 & 0\\
0 & 1 & 0 & 0
\end{array}\right).
\]

\[
A^{(8)}=P_{34}A^{(7)}P_{34}=\left(\begin{array}{cccc}
0 & 0.75 & 2.5 & 0.75\\
-0.75 & 0 & 1.75 & 0\\
-2.5 & -1.75 & 0 & -1.75\\
-0.75 & 0 & 1.75 & 0
\end{array}\right).
\]

\[
A^{(9)}=\textnormal{EQ}(A^{(8)},1,2)=\left(\begin{array}{cccc}
0 & 0 & 2.125 & 0.375\\
0 & 0 & 2.125 & 0.375\\
-2.125 & -2.125 & 0 & -1.75\\
-0.375 & -0.375 & 1.75 & 0
\end{array}\right).
\]

\[
A^{(10)}=P_{34}A^{(9)}P_{34}=\left(\begin{array}{cccc}
0 & -1.75 & -2.125 & -2.125\\
1.75 & 0 & -0.375 & -0.375\\
2.125 & 0.375 & 0 & 0\\
2.125 & 0.375 & 0 & 0
\end{array}\right).
\]

The ranking vector for $A^{(10)}$ is 
\[
w=(-1.5,0.25,0.625,0.625)^{T},
\]
so the final value of $q$ is 3. The weights of alternatives $p$
and $q$ are now equal and the highest, so the algorithm breaks. The
output matrix is $A^{10)}$.

Notice that the chosen alternative is not a sole leader in the ranking.
However, even the slightest correction of the element $a_{pq}$ in
favor of the alternative $p$ may change that. For example, if we
put $a_{34}=-0.1$ (and, respectively $a_{43}=0.1$), then we get
"the winning ranking": 
\[
w=(-1.5,0.25,0.6,0.65)^{T}.
\]
\end{example}

\subsubsection{The bubble algorithm\protect\label{subsec:The-bubble-algorithm}}

As Example \ref{greedy} shows, the greedy algorithm has some disadvantages.
It is fast on average, however, if the preferred alternative is on
the bottom of the ranking we may need to run a loop $n-1$ times.
Secondly, the whole procedure may competely reverse the ranking, which
is undesirable.

Therefore, we suggest an alternative algorithm, which promotes a chosen
alternative stepwise.

INPUT: $A,\ p$. 
\begin{enumerate}
\item $q:=$ the number of the best alternative 
\item If $p=q$ then return($A$); 
\item Construct the basis (\ref{basis}); 
\item Apply the Gram-Shmidt process to obtain the basis (\ref{ort-basis}); 
\item repeat\\

\begin{itemize}
\item $A:=$EQ($A,p,q$); 
\item $q:=$ the number of the alternative directly ahead of the alternative
$p$ in the ranking; 
\end{itemize}
until ranking($p$) $=$ ranking($q$);\\

\item return($A$); 
\end{enumerate}
\begin{example}
Consider once more the matrix $A$ from the example \ref{greedy}.
The output matrix after running the bubble algorithm is of the form:

\[
A^{(10)}=\left(\begin{array}{cccc}
0 & 1.625 & 3.875 & 0\\
-1.625 & 0 & 2.25 & -1.625\\
-3.875 & -2.25 & 0 & -3.875\\
0 & 1.625 & 3.875 & 0
\end{array}\right),
\]

and the final ranking vector is 
\[
w=(1.375,-0.25,-2.5,1.375)^{T},
\]

so the fourth alternative moved up to the first position (ex aequo
with the first one), but the relative positions of the other alternatives
did not change. 
\end{example}

\section{Monte Carlo Simulation\protect\label{sec:Monte-Carlo-Simulation}}

For Monte Carlo testing, we generated $2500$ preference profiles
within which the relative priority of a pair of alternatives ranges
from $[1/9,9]$. The number of alternatives ranges from $5$ to $9$,
i.e. for five alternatives we generate $500$ random profiles, for
6 alternatives - $500$ profiles were prepared, etc.

Based on the drawn preference profiles, we created random pairwise
comparison matrices (PCM) in such a way that for a preference profile
\[
w=\left(w(a_{1}),\ldots,w(a_{n})\right)^{T}
\]

a $C_{\alpha}$ is a $n\times n$ PCM in the form 
\[
C_{\alpha}=\left(\begin{array}{ccccc}
1 & c_{1,2}r_{1,2} & c_{1,3}r_{1,3} & \cdots & c_{1,n}r_{1,n}\\
c_{2,1}r_{2,1} & 1 & c_{2,3}r_{2,3} & \cdots & c_{2,n}r_{2,n}\\
\vdots & \cdots & \ddots & \cdots & \vdots\\
\vdots & \cdots & \cdots & \ddots & \vdots\\
c_{n,1}r_{n,1} & c_{n,2}r_{n,2} & \cdots & c_{n,n-1}r_{n,n-1} & 1
\end{array}\right),
\]
where 
\[
c_{ij}=\frac{w(a_{i})}{w(a_{j})},
\]
and $r_{ij}$ is a real number randomly selected from $[1/\alpha,\alpha]$
for $i,j=1,\ldots,n$. Thus, by increasing the value of $\alpha$,
we effectively increase the inconsistency of $C_{\alpha}$. We created
matrices in the form of $C_{\alpha}$ for all $2,500$ random preference
profiles and for all $\alpha$ values from the set $\{1,1.1,1.2,\ldots,4.9,5\}$.
In the end, we generated $102,500=2,500\times41$ random PCM matrices
with varying degrees of inconsistency and dimensions ranging from
$5\times5$ to $9\times9$. All generated matrices were used as input
to both greedy (Sec. \ref{subsec:The-greedy-algorithm}) and bubble
algorithms (Sec. \ref{subsec:The-bubble-algorithm}). 
\begin{figure}[h]
\subfloat[Greedy algorithm, LBN strategy]{\begin{centering}
\includegraphics[width=0.48\textwidth]{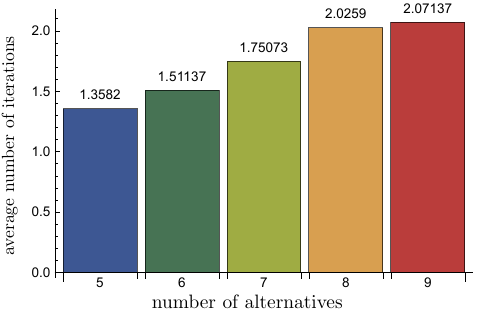}
\par\end{centering}
}~~\subfloat[Greedy algorithm, LBR strategy]{\begin{centering}
\includegraphics[width=0.48\textwidth]{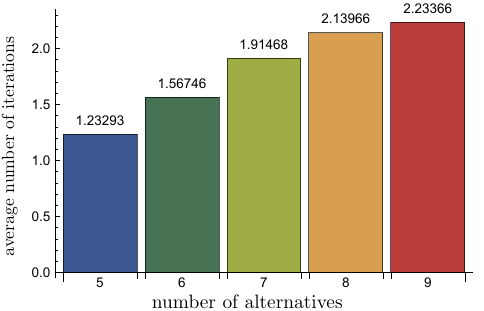}
\par\end{centering}
}

\subfloat[Bubble algorithm, LBN strategy]{\begin{centering}
\includegraphics[width=0.48\textwidth]{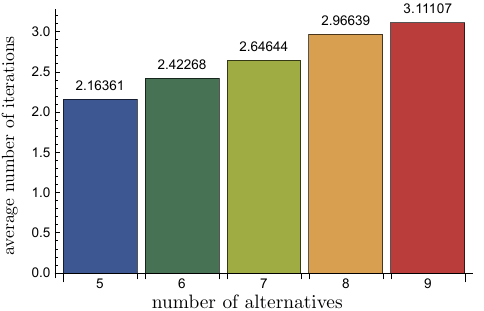}
\par\end{centering}
}~~\subfloat[Bubble algorithm, LBR strategy]{\begin{centering}
\includegraphics[width=0.48\textwidth]{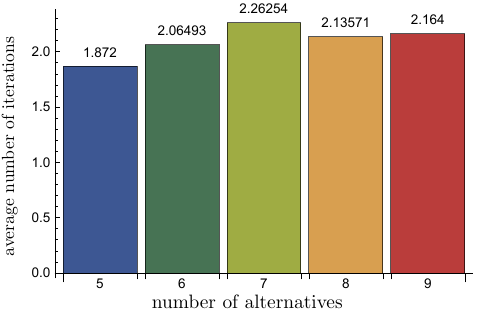}
\par\end{centering}
\label{fig:no-of-iterations-vs-no-of-alternatives-bubble-lbr}

}

\caption{Number of iterations vs. number of alternatives}

\label{fig:no-of-iterations-vs-no-of-alternatives}
\end{figure}

For both algorithms, we examined for two different strategies for
selecting the promoted alternative. In the first case, we took as
the subject of promotion the alternative with the n-th index (the
last in the sense of numbering) regardless of its actual ranking position.
In the second strategy, we first calculated the ranking using GMM
and then promoted the last alternative in the ranking. The first strategy
was called LBN - \textquotedbl last by numbering\textquotedbl{} and
the second LBR - \textquotedbl last by ranking.\textquotedbl{} Hence,
it took us $102,500$ $\times$ ($2$ algorithms) $\times$ ($2$
strategies) $=$ $410,000$ runs of the greedy and bubble algorithms
to conduct the assumed experiments.

\begin{figure}[H]
\begin{centering}
\includegraphics[width=0.8\textwidth]{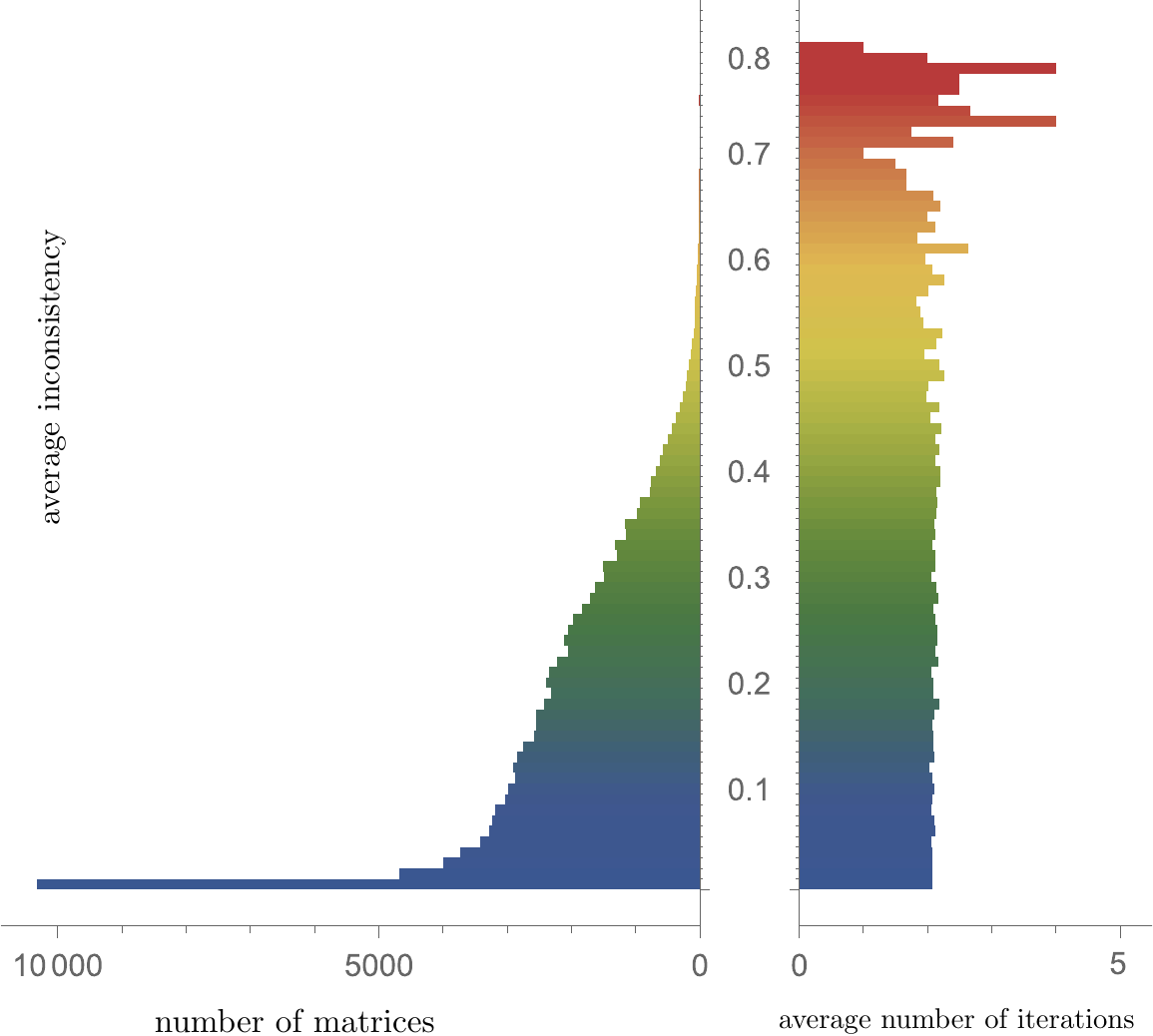}
\par\end{centering}
\caption{Number of iterations and number of tested matrices vs. inconsistency
using the bubble algorithm and LBR strategy as example.}

\label{fig:no-of-iter-vs-avg-inconsistency-bubble-lbr-all}
\end{figure}

In all four cases, the average number of iterations depends on the
number of alternatives (Fig. \ref{fig:no-of-iterations-vs-no-of-alternatives}).
In most cases, it increases as the number of alternatives increases.
The only exception was seen in the case of the bubble algorithm and
the LBR strategy where a greater number of alternatives does not necessarily
translate into an increased number of iterations (Fig. \ref{fig:no-of-iterations-vs-no-of-alternatives-bubble-lbr}).

While the relationship between the number of alternatives and the
number of iterations of the algorithms seems significant, there is
no evident relationship between the inconsistency of the tested matrices
and the number of iterations. In order to observe this possible relationship,
we divided the set of tested matrices into subsets where the first
one contained C matrices with CI(C) between 0 and $0.01$, the second
one between $0.01$ and $0.02$, and so on. For each interval, we
counted the average inconsistency, the average number of iterations
and the set count. As long as the set size did not fall below a few
tens of elements, the average number of iterations remained similar
regardless of the average inconsistency of the matrix in a given subset
(Fig. \ref{fig:no-of-iter-vs-avg-inconsistency-bubble-lbr-all}).
Since the result was similar in each of the four variants considered
in the figure, we used in (Fig. \ref{fig:no-of-iter-vs-avg-inconsistency-bubble-lbr-all})
the result for the bubble algorithm and the LBR strategy. It is worth
noting that the modifications made by the algorithm to the matrix
do not change its level of inconsistency. Thus, attempts to detect
such manipulation using only inconsistency measurements may be ineffective.

\begin{figure}
\begin{centering}
\includegraphics[width=0.48\textwidth]{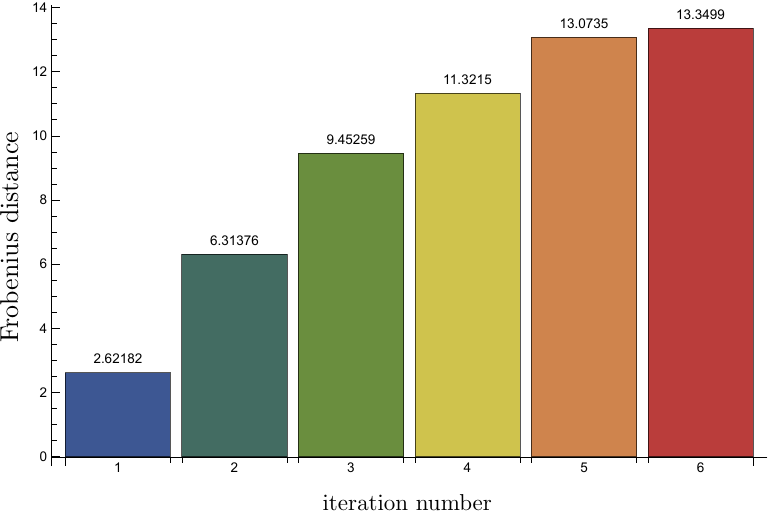}
\par\end{centering}
\caption{Frobenius distance between input matrix and its subsequent improvements.
Average values for greedy algorithm and LBN strategy.}

\label{fig:frobenius-distance-increase-greedy-lbn}
\end{figure}

The Frobenius distance between the input matrix and the matrices that
are the output of successive algorithms' iterations increases. This
is because each iteration changes subsequent elements of the matrix,
moving it away from the original matrix (Fig. \ref{fig:frobenius-distance-increase-greedy-lbn}).
This behavior can be observed regardless of the type of algorithm
and strategy adopted.

\begin{figure}
\begin{centering}
\includegraphics[width=0.48\textwidth]{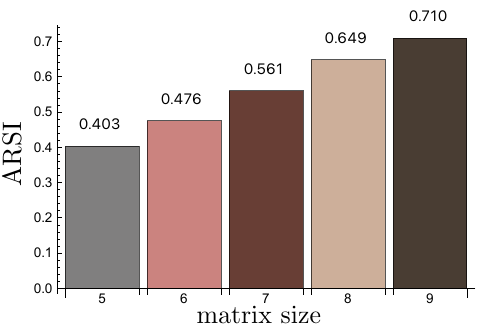}
\par\end{centering}
\caption{The average RSI (ARSI) value for the matrices studied depending on
the size of the matrix.}
\label{fig:arsi-vs-matrix-size}
\end{figure}

Similarly, a consistently observable pattern is the decline in the
Average Ranking Stability Index $\textit{ARSI}$ values. The $\textit{ARSI}$
values depend on the size of the matrix i.e., the larger the dimension
of the matrix, the higher the ARSI (Fig. \ref{fig:arsi-vs-matrix-size}).

\begin{figure}
\begin{centering}
\subfloat[PC matrices $5\times5$]{\begin{centering}
\includegraphics[width=0.48\textwidth]{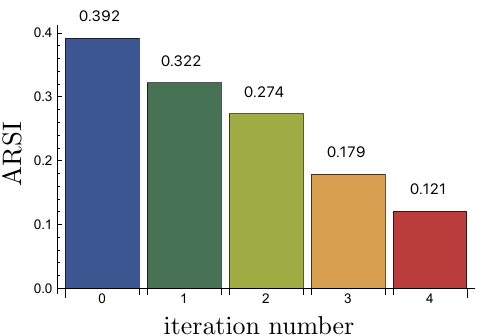}
\par\end{centering}
}~~\subfloat[PC matrices $9\times9$]{\begin{centering}
\includegraphics[width=0.48\textwidth]{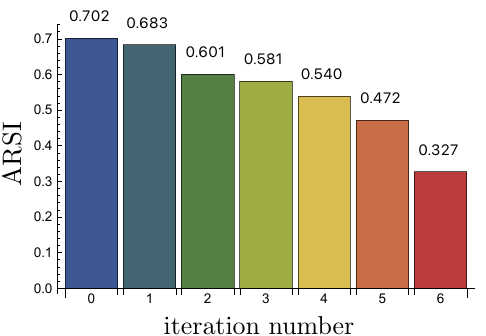}
\par\end{centering}
}
\par\end{centering}
\caption{Decreasing (average) value of $\textit{ARSI}$ for modified matrices
in successive iterations using the greedy algorithm and LBN strategy
as an example. Iteration $0$ shows the average $\textit{ARSI}$ value
for the unmodified input matrices.}
\label{fig:decreasing-arsi-example}
\end{figure}

Therefore in the study we calculated the corresponding values in groups
of matrices of the same dimensions (Fig. \ref{fig:decreasing-arsi-example}).
This corresponds to the intuitive observation that making the first
intervention is the most difficult. Each subsequent one comes more
and more easily. More formally, $\textit{ARSI}$ is being reduced
in subsequent iterations of the algorithm, since they make two alternatives'
weights equal and closer to the rest and leave other alternatives'
weights unchanged. This implies that each manipulation increases the
possibility of other manipulations.

\section{Conclusions}

In the presented work we have introduced two algorithms of promoting
a given alternative to the position of a ranking leader. They are
both based on the EQ algorithm equating two given alternatives in
a ranking. The first one, called the greedy algorithm, in each step
equates the rankings of a promoted alternative and the current leader.
The second one (the bubble algorithm) in each step equates an alternative
with the one directly preceding it in the ranking. We have also defined
the Average Ranking Stability Index (ARSI) for a PC matrix to measure
how easily the data manipulation may happen.

The Monte Carlo study has shown that in general it is harder to create
a new leader when there are more alternatives. On the other hand,
the input inconsistency of data has no influence on the ease of manipulation.
The third conclusion is that each each manipulation facilitates the
subsequent ones. The final remark is that the EQ algorithm does not
change the scale, i.e. if the input PC matrix elements have been taken
from the range $[-M,M]$, the output matrix elements had the same
property.

\section{Acknowledgments}

The research has been supported by The National Science Centre, Poland,
project no. 2021/41/B/HS4/03475 and by the Polish Ministry of Science
and Higher Education (task no. 11.11.420.004).


\bibliographystyle{plain}
\addcontentsline{toc}{section}{\refname}\bibliography{papers_biblio_reviewed}

\end{document}